\documentclass{ecai}
\usepackage{graphicx}
\usepackage{latexsym}
\usepackage{url}
\usepackage{grbbridge}
\usepackage{todonotes} 
\usepackage{float}
\usepackage{dblfloatfix}

\begin{document}
\begin{frontmatter}
\title{$\mathbf{BridgeHand2Vec}$ Bridge Hand Representation}

\author[A]{\fnms{Anna}~\snm{Sztyber-Betley}\orcid{0000-0002-6464-8194}\thanks{Corresponding Author. Email: anna.sztyber@pw.edu.pl}}
\author{\fnms{Filip}~\snm{Kołodziej}}
\author{\fnms{Jan}~\snm{Betley}\orcid{0009-0008-3518-191X}} 
\author[A]{\fnms{Piotr}~\snm{Duszak}\orcid{0000-0003-0828-1727}}

\address[A]{Warsaw University of Technology}

\begin{abstract}
Contract bridge is a game characterized by incomplete information, posing an exciting challenge for artificial intelligence methods. This paper proposes the $BridgeHand2Vec$ approach, which leverages a neural network to embed a bridge player's hand (consisting of 13 cards) into a vector space. The resulting representation reflects the strength of the hand in the game and enables interpretable distances to be determined between different hands. This representation is derived by training a neural network to estimate the number of tricks that a pair of players can take. In the remainder of this paper, we analyze the properties of the resulting vector space and provide examples of its application in reinforcement learning, and opening bid classification. Although this was not our main goal, the neural network used for the vectorization achieves SOTA results on the DDBP2 problem (estimating the number of tricks for two given hands).
\end{abstract}
\end{frontmatter}

\section{INTRODUCTION}

Artificial intelligence methods in games are currently of great interest to researchers, with impressive results in Perfect Information Games (PG) like chess or Go \cite{Silver2016,Silver2017}. Strategies in Imperfect Information Games (IG) are more challenging. In poker, winning solutions have recently been obtained against top players \cite{Brown2017,Moravcik2017}. Bridge (and similar trick-taking games like Spades \cite{Cohensius2019, baier2018}) is still challenging. In addition to incomplete information, the challenge in the bridge is the cooperation between partners.

Four people participate in a game of bridge; pairs of players form cooperative teams. The game is played with a deck of 52 cards. Each player has 13 cards called a hand. A player knows his cards but does not know the other players' cards, which constitutes confidential information. The game consists of two phases: auction and gameplay. The auction leads to the determination of a trump suit and the contract level. During bidding, partners must communicate with each other based on a limited set of bids. The gameplay is based on a collection of tricks. In terms of gameplay, effective algorithmic solutions use advanced search techniques with heuristics for speed up. In terms of bidding, on the other hand, programs have still not reached the level of professional players. Bridge bots play The World Computer-Bridge Championship. The last championship was held in 2019 and was won by Micro Bridge\footnote{\url{https://micro-bridge.software.informer.com/}}, beating Synrey\footnote{\url{http://www.synrey.com}} in the final. The previous three editions were won by WBridge5\footnote{\url{http://www.wbridge5.com/}} \cite{Ventos2017}. Leading bridge bidding programs use rigid, predefined rules based on human bidding systems.

\begin{figure}
\centerline{\includegraphics[scale=1]{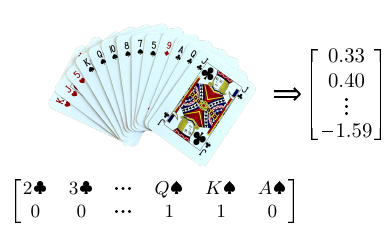}}
\caption{Bridge hand, binary representation (bottom), and vectorized $BridgeHand2Vec$ representation (right). } \label{fig:hand_vector}
\end{figure}

In this study, we introduce the $BridgeHand2Vec$ approach for acquiring a vector-based representation of a bridge hand that captures the similarity of hands based on their strength in the game (see Fig.~\ref{fig:hand_vector}). Our method involves training a neural network to replicate the outcomes of the Double Dummy Solver (DDS). The DDS is a tool that provides information on the number of tricks each player should take when playing in a given trump suit, with the assumption that all cards are exposed and all players play optimally.

$BridgeHand2Vec$ representation has the following applications:
\begin{itemize}
\item it allows for the determination of interpretable distances between hands and the search for similar hands;
\item accelerates the performance of learning algorithms;
\item increases the sample efficiency of learning algorithms. This is an issue relevant for bridge bot-human collaboration, where acquiring large amounts of data, e.g. to teach a non-standard bidding system, can be problematic.
\item Many learning algorithms in bridge bidding use (or attempt to use) partner's hand estimation \cite{Zhang2020,Zhang2022,Rong2019,Gong2020,Tian2020}. $BridgeHand2Vec$ provides better cost functions and methods for assessing estimation quality. 
\item Bridge players are obligated to disclose their system arrangements to their opponents. Currently, bridge programs use rule sets. However, experiments using reinforcement learning \cite{Yeh2016,Rong2019,Lockhart2020,Gong2020,Tian2020,Tian2020a} indicate that it is possible to improve here by creating a custom system, but this poses problems with revealing arrangements. $BridgeHand2Vec$ can provide a tool for revealing agreements or sampling representative cards for a bidding sequence.
\item $BridgeHand2Vec$ can also provide an analytic tool for players. For example, we want to determine the correct bid with a given hand. To do this, we can calculate the distances between a "perfect" hand for each possible bid and the problematic hand and select the nearest one.
\end{itemize}  
The proposed approach of meaningful representation of hidden states can be generalized to other games and algorithms with hidden state estimation or belief modelling (e.g. Policy Belief Learning \cite{Tian2020}). Hand representation can also be used for collectible card video games like Hearthstone \cite{Janusz2018}. 

The rest of the paper is structured as follows. Section~\ref{sec:bridge} introduces bridge rules and strategy. Section~\ref{sec:related_work} presents related work. Section~\ref{sec:bridgehand2vec} introduces $BridgeHand2Vec$ algorithm, compares it with existing results for DDS estimation, and explores properties of resulting vector space. Section~\ref{sec:applications} presents exemplary applications, and Section~\ref{sec:conclusion} concludes the paper.

\subsection{Bridge rules}
\label{sec:bridge}

\begin{figure}
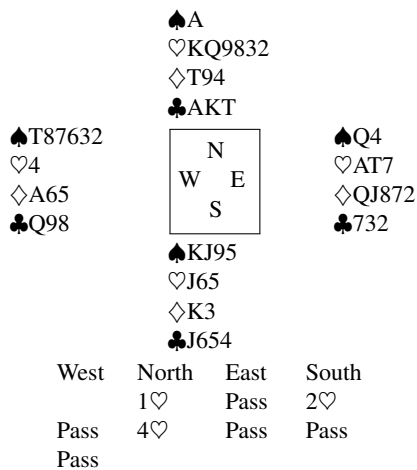

\centering
\begin{handdiagram}
\north{A,KQ9832,T94,AKT}
\south{KJ95,J65,K3,J654}
\east{Q4,AT7,QJ872,732}
\west{T87632,4,A65,Q98}
\end{handdiagram}
\auction{,1h,p,2h,p,4h}
\caption{Exemplary board and bidding.} 
\label{fig:ex_board}
\end{figure}

An example board is shown in Fig.~\ref{fig:ex_board}. The game is played with a full deck of 52 cards distributed among four players. The players are denoted as North (N), South (S), East (E), and West (W). NS and EW form pairs. The game consists of two phases: auction and gameplay (tricks taking).

An auction example is illustrated at the bottom of Fig.~\ref{fig:ex_board}. The auction starts with the dealer (N in this example). During the auction, players can either bid a pass or a call, comprising a level and a denomination, which can be a trump suit or no trump. For brevity, the description excludes double and redouble. For instance, the 1$\spadesuit$ bid denotes a declaration to win seven tricks with spades as the trump suit. The auction is won by the pair with the highest bid, and the person who first declares the final contract's denomination is the declarer. In the given example, the final contract is 4$\heartsuit$, requiring ten tricks to be won. The NS pair wins the bidding, and N becomes the declarer. Subsequently, the game advances to the trick-taking phase.

The person to the declarer's left (E in the example) plays the first card. The declarer's partner (S) puts his cards on the table as the dummy. The trick is formed by four cards played clockwise by the following players. There is an obligation to follow the suit of the first card in the trick. The player who played the highest card takes the trick. This player starts the next trick. If a trump suit is played, the highest card of this suit wins the trick. The game continues until all cards are played. If the contract is completed (the declared number of tricks or more is taken), the points go to the pair playing the hand. Otherwise, the points go to the defenders. Some contracts (e.g. 4$\spadesuit$) give an additional scoring bonus.

During auction, players use specific heuristics to evaluate a hand. One of the more popular heuristics is High Card Points (HCP), calculated as four points for each ace, three for a king, two for a queen and one for a jack. The N hand, for example, has 16 HCP. Players also use a predetermined bidding system. For example, the common meaning of 1$\heartsuit$ opening is "at least 5 hearts and at least 12 HCP".

Table~\ref{tab:dds_ex_results} shows an instance of DDS results, which indicate the number of tricks players are expected to win in each trump suit (or no trump). The results were generated with the Bridge Calculator\footnote{http://bcalc.w8.pl/} \cite{Beling2017}. Other popular solvers include GIB \cite{Ginsberg2001} and Bo Haglund's DDS\footnote{https://github.com/dds-bridge/dds}.

\begin{table}
\centering
\caption{Exemplary results of DDS}
\label{tab:dds_ex_results}
 \begin{tabular}{c|ccccc}
declarer & $\clubsuit$ & $\diamondsuit$ & $\heartsuit$ & $\spadesuit$ & NT \\\hline
  N & 9 & 6 & 10 & 7 & 8\\
  S & 9 & 6 & 10 & 7 & 8\\
  E & 3 & 7 & 3 & 6 & 3\\
  W & 4 & 7 & 3 & 6 & 3\\
\end{tabular}
\end{table}

In the reminder of the paper the following notation will be used. The hands are presented in PBN (Portable Bridge Notation). A hand is represented by a sequence of 13 cards in suit and rank order. The sequence starts with the cards in the spade suit, followed by the hearts, diamonds, and clubs. Suits are separated by dots. T denotes ten. N hand in the example (Fig.~\ref{fig:ex_board}) is represented in PBN as A.KQ9832.T94.AKT. The hand distribution is represented by four integers and describes the length of the suits in the decreasing length order, in the example N hand has the distribution 6331.

\subsection{Related work}
\label{sec:related_work}

Although current state-of-the-art bridge programs can compete with advanced bridge players in terms of gameplay, they fall short in bidding. One of the potential reasons for this is the employment of rigid bidding rules. Consequently, reinforcement learning in bidding has become a significant area of interest in AI-based bridge solutions. Additionally, experiments on estimating the number of tricks to be taken in a game using neural networks have recently emerged.

Reinforcement learning in bridge bidding is a fascinating problem because a pair of players can communicate using their own "language" - a bidding system. Researchers have employed two approaches: learning from scratch or pretraining using human bidding systems. The first deep reinforcement learning approach was proposed in the work \cite{Yeh2016} and was limited to noncompetitive bidding scenarios, where it is assumed that opponents pass all the time. More advanced solutions adapted to competitive bidding were subsequently presented in works \cite{Rong2019,Lockhart2020,Gong2020,Tian2020,Tian2020a}. Additionally, \cite{Zhang2020} presented an LSTM network for response classification learned from historical boards. The other papers used reinforcement learning algorithms: \cite{Rong2019} uses two networks, a policy network and a hand estimation network; \cite{Lockhart2020} proposes a solution tailored to play with humans pre-trained on WBridge5 bots; \cite{Gong2020} proposes a solution without expert knowledge and modelling of partner's card; \cite{Tian2020} proposes a PBL (Policy Belief Learning) algorithm, and \cite{Tian2020a} presents a Joint Policy Search (JPS) algorithm. These solutions are often compared with one of the leading bridge programs, WBridge5, and the authors report that the learned system wins in bidding. The best results were reported in the paper \cite{Tian2020a}.
However, to the best of the authors' knowledge, no commercial programs currently utilize deep learning in bidding. One limitation in this regard may be the requirement to reveal the bidding system to opponents, which is challenging for black-box models. The first work aimed at disclosing the bidding system was presented in \cite{Zhang2022}.

The concept of estimating the partner's hand is prevalent in research on bidding learning. However, it is sometimes pointed out that such estimation does not improve the overall quality of the algorithm \cite{Gong2020}. A probability distribution for all 52 cards in the deck is usually used to represent the partner's hand. This representation does not consider the specifics of bridge play and the relevance of individual cards. While after a long auction, professional players are usually able to estimate the overall strength and shape of the partner's hand (e.g. 12-14 HCP, five hearts in a 5332 distribution) or even the position of key figures (what is the chance that the partner has the A$\heartsuit$), estimating with reasonable accuracy whether a partner has a specific low card (e.g. the 2$\clubsuit$) is never possible. Therefore, the authors believe the cost functions used in these works are ineffective. In this work, we propose a vectorization $BridgeHand2Vec$, which allows the determination of the Euclidean distance between hands that reflects well the strength in a bridge game.

The second issue investigated is learning the number of tricks possible to be taken in the play phase, i.e. reproducing the results of a DDS solver. In the paper \cite{Mossakowski2009}, the authors formulate two Double Dummy Bridge Problem (DDBP) type problems: DDBP2 - where only two hands are known, and DDBP4 - where all four hands are known. Various works have analyzed this problem, including \cite{Kowalik2021,Mandziuk2019,Mernagh2016,Mossakowski2009}. In this study, we train a neural network to solve the DDBP2 problem to obtain a vector representation, $BridgeHand2Vec$, of the hand. The results obtained in this study are competitive with those reported in the literature and, for some types of problems, better than those obtained by professional bridge players who operate under time pressure. Detailed results are presented in Section \ref{sec:dds_train}. Neural networks solving DDBP problems cannot achieve the accuracy of search-based DDS solvers, but they are faster.

\section{$\mathbf{BridgeHand2Vec}$ representation}
\label{sec:bridgehand2vec}
\subsection{Neural network training for DDBP}
\label{sec:dds_train}

In order to obtain a vector representation of a hand, a neural network was trained to reproduce DDS results. The DDS algorithm determines the number of tricks the declarer can take in each suit and no-trump, assuming complete information and optimal plays by all players. The number of tricks determined by the DDS differs slightly from the results achieved in a real game (see discussion and numerical experiments in \cite{Rong2019}) but is a useful approximation.

The training data were generated using Bridge Calculator \cite{Beling2017}. The purpose of the network is to estimate the average number of tricks based on the hands of the two players, N and S (given an unknown distribution of the EW hands). This problem is referred to as DDBP2 in the work of \cite{Mossakowski2009}. The outcome of the board also depends on the position of the cards in the EW hands, so for each NS hand, 10 EW hands were generated, and the results obtained from the DDS were averaged. (A similar approach was used in the work of \cite{Tian2020} to estimate the outcome of a board by averaging 20 opponents' hands.) For training, 400,000 hands were generated. Each hand was also flipped (swapping the N and S hands), resulting in 800 000 training examples. An exemplary row of training data was shown in Table~\ref{tab:training}. N $\clubsuit$ denotes the number of tricks to be taken on average in clubs when N is declarer. It can be noted that changing the declarer has a slight impact on the number of tricks.

\begin{table*}
\centering
\caption{Exemplary row of training data}
\label{tab:training}
 \begin{tabular}{c|c|c|c|c|c|c|c|c|c|c|c}
hand N & hand S & N $\clubsuit$ & N $\diamondsuit$ & N $\heartsuit$ & N $\spadesuit$ & N NT & S $\clubsuit$ & S $\diamondsuit$ & S $\heartsuit$ & S $\spadesuit$ & S NT\\\hline
A72.K7.AQJ74.A94 & J.J852.T9.KQ6532 & 11.1 & 10.5 & 7.3 & 5.0 & 10.6 & 11.1 & 10.6 & 7.2 & 4.9 & 10.6\\
\end{tabular}
\end{table*}

For the first version of the network, an asymmetry was noted for estimating the number of tricks in different suits. For example, for hand N A432.A432.432.32 and hand S KQJ5.KQJ5.76.765, the model predicted 8.25 tricks in hearts and 8.83 tricks in spades, despite the identical cards in these suits.
Therefore, suit rotation was applied: K2.A76543.Q2.J32 $\rightarrow$ A76543.Q2.J32.K2 $\rightarrow$ J32.K2.A76543.Q2 $\rightarrow$ Q2.J32.K2.A76543, giving three additional training examples from each original example, without the need for DDS calculation (partner's hand and trick numbers were rotated accordingly). The model trained on data including rotation gives, for the hands above, 8.1499 tricks in hearts and 8.1407 tricks in spades, a much smaller disparity. Finally, the set of 3 200 000 examples was used for training.

\begin{figure}
\centerline{\includegraphics[scale=0.65]{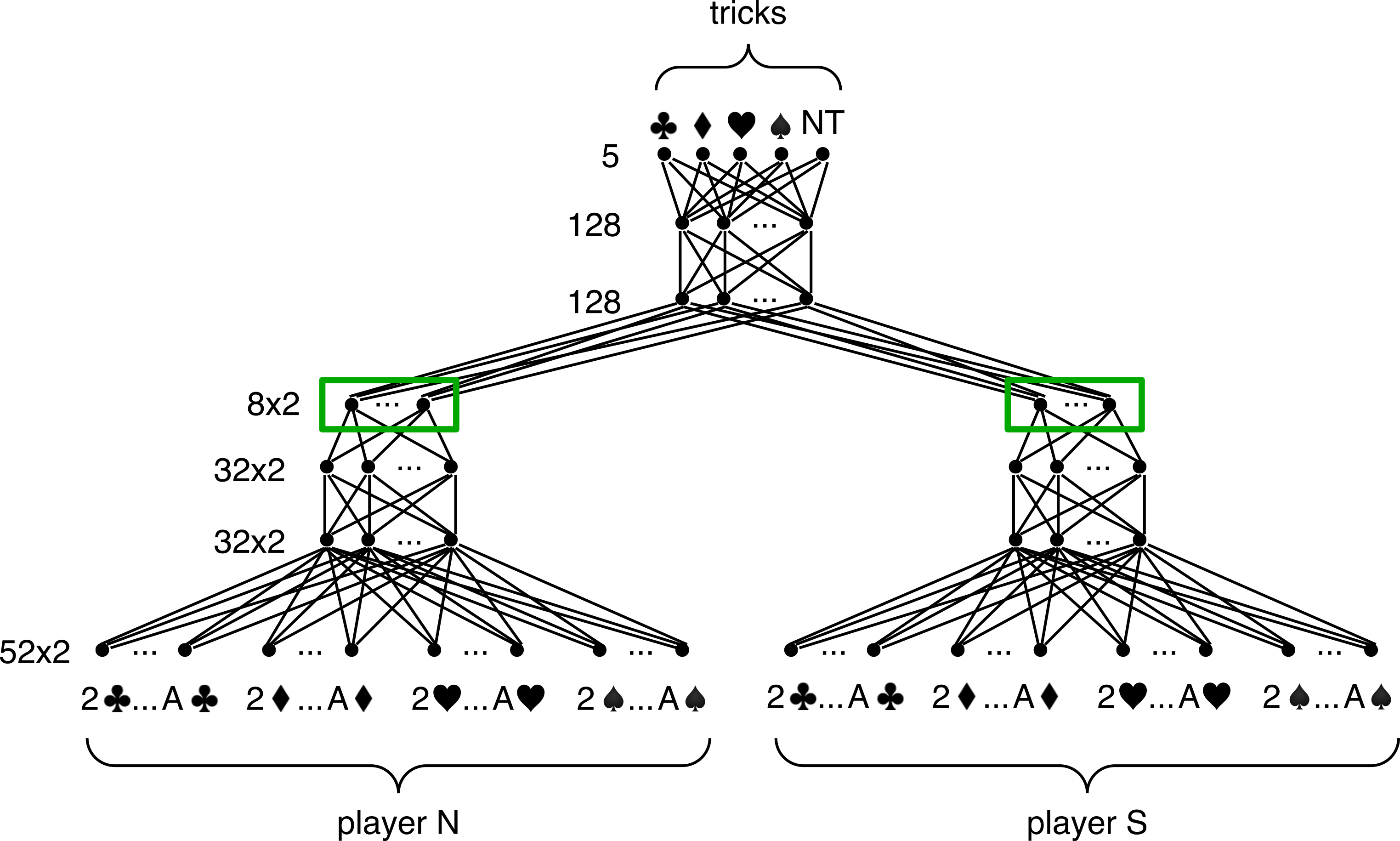}}
\caption{$BridgeHand2Vec$ network structure. The embedding layer was marked by green rectangles.} \label{fig:nn}
\end{figure}

The network structure is shown in Fig.~\ref{fig:nn}. Hands N and S are represented at the input as binary vectors of length 52. Player N is always the declarer (to change the declarer, the order of the network's inputs has to be swapped). Then the input is subjected to vector embedding (two dense layers with 32 neurons), and at the output, we get a vector with 8 elements representing the hand - $BridgeHand2Vec$ embedding. This layer includes batch normalisation without affine transform to get vectors with normalised components. The critical point is that the embedding layers for both hands share weights. Also, the batch normalisation layers share coefficients. The vector representations are then concatenated and fed into two dense layers with 128 neurons. The output of the network is five neurons with linear activation. The loss function is the mean squared error. The ELU activation function was applied to all hidden layers. The Adam optimizer was used for training.

\begin{figure}
\centerline{\includegraphics[scale=0.7]{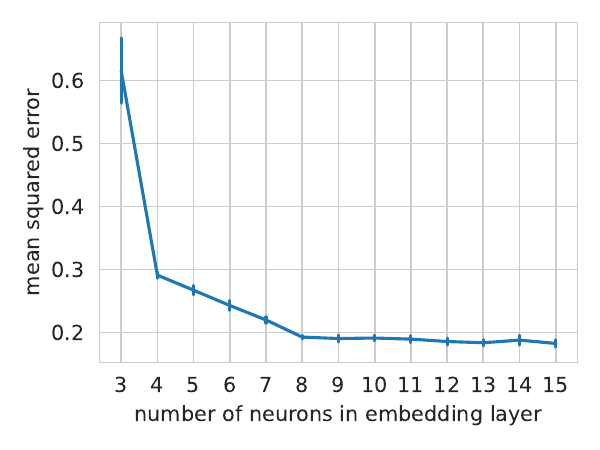}}
\caption{Mean squared error depending on vector representation size.} 
\label{fig:vec_size}
\end{figure}

The vector representation size was chosen based on the results on the validation set (50 000 hands). The values of mean squared errors averaged over the ten runs are shown in Fig.~\ref {fig:vec_size}. The error bars represent the standard deviation. It can be observed that when increasing the representation size beyond 8, the estimation accuracy increases only slightly.

The accuracy of the estimate of the number of tricks is shown in Table~\ref{tab:dds_results}. The following columns show the percentage of hands from the test set (200 000 boards) for which the estimation error is less than 0.5, 1 and 2, respectively. For this table, both the prediction and the target value are real numbers. Ten models were trained to estimate the variability of the results, mean denotes average accuracy, and std denotes standard deviation.

\begin{table}
\centering
\caption{$BridgeHand2Vec$ trick error (not rounded).}
\label{tab:dds_results}
 \begin{tabular}{|c||c|c|c|}
	\hline
	Trick error & 0.5 & 1 & 2 \\
	\hline
	Suit mean & $83.952\%$ & $99.16\%$ & $99.996\%$ \\
 	Suit std & $0.487\%$ & $0.051\%$ & $0.0003\%$ \\
	\hline
	No-trump mean & $72.21\%$ & $94.86\%$ & $99.78\%$ \\
    No-trump std & $0.634\%$ & $0.225\%$ & $0.018\%$ \\
	\hline
\end{tabular}
\end{table}

The accuracy obtained for the problem of estimating DDS results using a neural network was compared with results from the literature. The main work in this area is \cite{Mossakowski2009}, and the presentation of the results has been adapted to the conventions of this paper. The summary also includes the results obtained in \cite{Mandziuk2018} with an autoencoder network and in  \cite{Kowalik2021} with a convolutional network. The results are presented in Table~\ref{tab:dds_results_round}. In the work~\cite{Mossakowski2009}, the network estimates a discrete number of tricks, so our results have been rounded. The consecutive columns of the table illustrate the percentage of examples for which the estimation is accurate (column 0) or the error is at most 1 or 2 tricks. DDBP2 is a problem where only NS cards are known. In DDBP4, all cards are known. $BridgeHand2Vec$ always works as DDBP2. The rows with the annotation 'human' contain the results obtained by the leading Polish players in the tests conducted in the paper \cite{Mossakowski2009}. Note that the human players had a limited time to solve the task. The results obtained by $BridgeHand2Vec$ are better than the best results obtained by single neural networks and are better than those obtained by human players under time pressure for suit contracts. Slightly better results than $BridgeHand2Vec$ are obtained by ensemble models for the no-trump contracts for the DDBP-4 problem. $BridgeHand2Vec$ solves the DDBP2 problem (it only knows the NS cards) and, in this respect, presents the best results compared to those described so far.

\begin{table}
\centering
\caption{Comparison of machine learning solutions for DDBP problems (predictions and target values rounded)}
\label{tab:dds_results_round}
 \begin{tabular}{|c|c|c|c|}
	\hline
	Trick error & 0 & 1 & 2 \\
	\hline
	Suit $BridgeHand2Vec$ mean & $71.33\%$ & $99.86\%$ & $99.999\%$ \\
    Suit $BridgeHand2Vec$ std & $0.314\%$ & $0.007\%$ & $0.0002\%$ \\\hline
	Suit MLP DDBP4 \cite{Mossakowski2009} & $37.80\%$ & $84.31\%$ & $97.34\%$ \\
	Suit MLP DDBP2 \cite{Mossakowski2009} & $40.13\%$ & $88.00\%$ & $98.77\%$ \\
	Suit AE-MLP DDBP4 \cite{Mandziuk2018} & $51.28\%$ & $95.33\%$ & $99.72\%$ \\
	Suit CNN DDBP4 \cite{Kowalik2021} & $58.42\%$ & $97.84\%$ & $99.89\%$ \\
	Suit CNN ensemble DDBP4 \cite{Kowalik2021} & $64.13\%$ & $95.39\%$ & $98.61\%$ \\
	Suit DDBP4 human \cite{Mossakowski2009} & $53.06\%$ & $81.63\%$ & $88.34\%$ \\
	Suit DDBP2 human \cite{Mossakowski2009} & $38.63\%$ & $81.20\%$ & $93.68\%$ \\
	\hline
	No-trump $BridgeHand2Vec$ mean & $63.30\%$ & $98.46\%$ & $99.92\%$ \\
    No-trump $BridgeHand2Vec$ std & $0.445\%$ & $0.077\%$ & $0.010\%$ \\\hline
	No-trump MLP DDBP4 \cite{Mossakowski2009} & $53.11\%$ & $96.48\%$ & $99.88\%$ \\
	No-trump MLP DDBP2 \cite{Mossakowski2009} & $34.66\%$ & $80.88\%$ & $96.07\%$ \\
	No-trump AE-MLP DDBP4 \cite{Mandziuk2018} & $41.73\%$ & $86.18\%$ & $96.63\%$ \\
	No-trump CNN DDBP4 \cite{Kowalik2021} & $57.24\%$ & $93.17\%$ & $98.03\%$ \\
	No-trump CNN ensemble DDBP4 \cite{Kowalik2021} & $63.83\%$ & $98.83\%$ & $99.94\%$ \\
	No-trump DDBP4 human \cite{Mossakowski2009} & $73.68\%$ & $88.30\%$ & $94.74\%$ \\
	No-trump DDBP2 human \cite{Mossakowski2009} & $43.32\%$ & $79.18\%$ & $93.17\%$ \\	
	\hline
\end{tabular}
\end{table}

A neural model designed to estimate the number of tricks improves SOTA compared to other solutions using neural networks (or other machine learning algorithms) and is also competitive with players under time pressure. However, the network results are not accurate enough to replace DDS in analyses, but the model is faster than solvers. Additionally, the $BridgeHand2Vec$ representation has exciting properties, described in the next section.

Code for model training and hand vectorization, along with the training data, can be found at \url{https://github.com/johny-b/BridgeHand2Vec}.

\subsection{$\mathbf{BridgeHand2Vec}$ properties}

This section will analyze the properties of the resulting vector space. The distribution of the vectors was examined on a test set of 200,000 hands. Due to the use of the batch normalisation layer, all vector components should have a mean of 0 and a standard deviation of 1. This is indeed the case. The respective means and standard deviations for the individual vector components are shown in Table~\ref{tab:mean_std}. 

\begin{table*}
\centering
\caption{Means and standard deviations of the vectors' components}
\label{tab:mean_std}
 \begin{tabular}{c|c|c|c|c|c|c|c|c}
& $v_0$ & $v_1$ & $v_2$ & $v_3$ & $v_4$ & $v_5$ & $v_6$ & $v_7$\\\hline
mean & 0.0063 & 0.0086 & 0.0093 & -0.0031 & -0.0117 & -0.0003 & 0.0043 & 0.0112\\
std & 0.9980 & 1.0011 & 1.0024 & 0.9992 & 1.0007 & 1.0006 & 0.9985 & 1.0046\\
\end{tabular}
\end{table*}
        
The resulting $BridgeHand2Vec$ vector representation allows the calculation of distances and hand similarities. It can be noted that the similarities thus obtained reflect the strength of the hand in the game and not the number of identical cards, as in the classical binary vector representation. The Euclidean distance is used as a metric.

All vector components are centred around zero, so the zero vector should correspond to the average hand. By searching for hands with vector representation closest to the zero vector, we obtain the following: A72.K432.JT9.Q96, QT6.A432.Q95.Q94, QT6.Q95.K432.A32, A62.K432.Q86.Q96. These hands have the most balanced distribution possible (4333) and a strength of 10 - 11 HCP. The entire deck has 40 HCP, so one player receives 10 HCP on average. The resulting hands can be considered a good approximation of the average hand.

Table~\ref{tab:neigh_hands} shows the closest hands to AKQ2.QJT987.32.2 together with the distances. The hands were searched from a set of 200,000 hands. It can be observed that the nearby hands have similar strength and composition, and some of the key cards (A$\spadesuit$ or Q$\heartsuit$) are repeated. However, there is no notable similarity in terms of small cards.

\begin{table}
\centering
\caption{Nearest neighbours of AKQ2.QJT987.32.2}
\label{tab:neigh_hands}
 \begin{tabular}{c|c}
hand & distance \\\hline
AKQ2.QJT987.32.2 & 0 (query hand)\\
AKJ6.QJ6532.T7.9 & 0.178\\
AQJT.KT9854.J4.7 & 0.215\\
AKQJ.KJ9864.85.6 & 0.232\\
AQJT.KJ9654.62.4 & 0.240\\
AKT5.KT6432.92.5 & 0.282\\
\end{tabular}
\end{table}

By analogy with the property of word embeddings \cite{Mikolov2013} (the famous $woman + (king - man)=queen$ equation), similar interesting properties of $BridgeHand2Vec$ space were explored. It turns out that when searching for vectors $x$ satisfying the equation $y + (H_{1}-H_{2}) \approx x$ for $H_{1}$ = A82.A7643.Q62.Q5, $H_{2}$ = A82.Q764.Q62.Q53, we obtain the cards shown in Table~\ref{tab:eq}. Thus, the addition of the vector $H_{1}-H_{2}$ can be interpreted as swapping the queen for an ace in hearts, adding a heart and taking a club. The results obtained for $y=$ 765.Q432.432.432 correspond to the obvious interpretation. Interesting results are obtained for $y=$ J6543.432.32.432. Here we get an interpretation of adding a card in hearts and an ace in the longest suit.

\begin{table}
\centering
\caption{Approximate solutions of $y + (H_{1}-H_{2})  \approx x$}
\label{tab:eq}
 \begin{tabular}{c|c}
$y$ & $x$ \\\hline
765.Q432.432.432 & 765.A6532.532.72\\
 & 752.A9542.532.32\\
 & T62.A8543.632.62\\ 
 & 865.A7432.962.94\\ 
 & T52.A6432.964.32\\\hline
J6543.432.32.432 & A7653.9875.63.32\\
 & A7532.7632.83.T2\\
 & A7652.T763.T3.86\\
 & A6432.J853.98.32\\
 & T9543.9842.74.98\\
\end{tabular}
\end{table}

\begin{figure}[t]
\centerline{\includegraphics[scale=0.7]{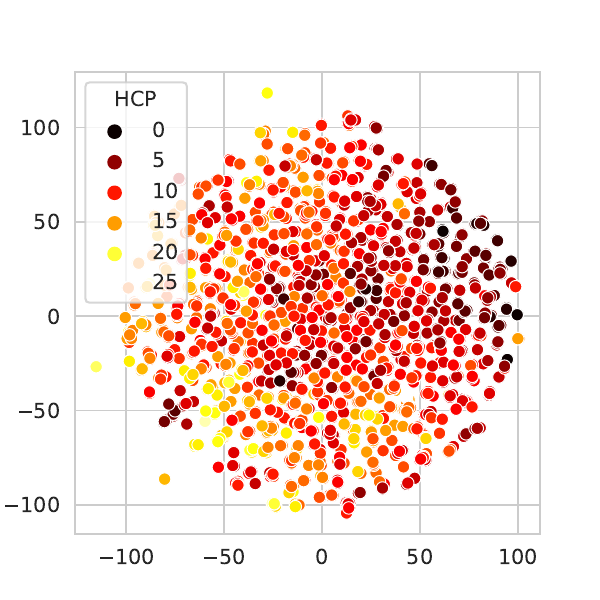}}
\caption{$BridgeHand2Vec$ embedded in 2D - colours correspond to the number of HCP}
\label{fig:tsne_hcp}
\end{figure}

A visualisation of the immersion of $BridgeHand2Vec$ space into two-dimensional space was carried out using the t-SNE algorithm \cite{Maaten2008}. In Fig.~\ref{fig:tsne_hcp}, the colours of the points correspond to the hand strength determined according to HCP.

\begin{figure}[t]
\centerline{\includegraphics[scale=0.7]{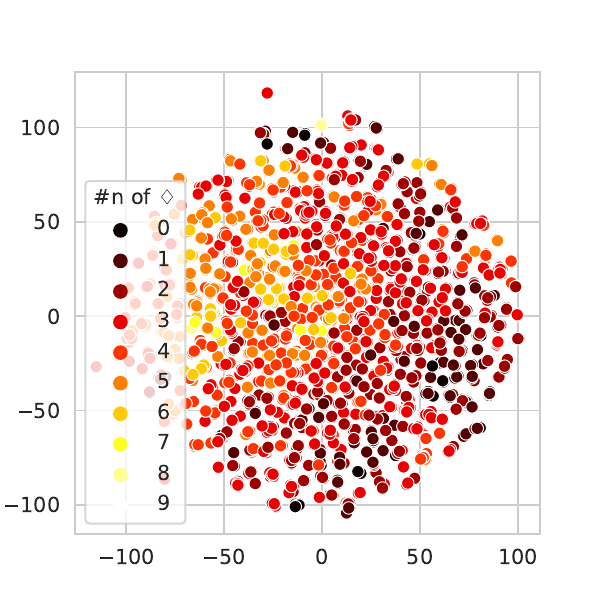}}
\caption{$BridgeHand2Vec$ embedded in 2D - colours correspond to the number of diamonds} 
\label{fig:tsne_suit}
\end{figure}

In Figure~\ref{fig:tsne_suit}, the points are coloured according to the number of diamonds in hand. Further analysis and visualisation suggest that the corresponding coordinates of the $BridgeHand2Vec$ vector are strongly correlated with hand strength (HCP) and the number of cards in each suit. 

\section{Applications}
\label{sec:applications}

The similarity between bridge hands is a central issue of the bidding process. In most use cases, "What is the correct bid given a hand in a given system?" can be restated as "We know some correct hands for all available bids. Which model hand is most similar to the given hand?". There is no known, obvious "similarity" metric because different hand traits must be taken into account (strength, length of suits, high cards) - hand evaluation is considered to be one of the essential skills in bridge\footnote{There are many books where this is the main topic, e.g. \cite{Bergen2002} and \cite{Lawrence1983}}, usually acquired with experience. Although bridge players use many different heuristics to assess hand value, there is a single goal: the more similar are two hands, the better the chance that the best bid with one will also be the best with the other. This translates directly to the metric based on the $BridgeHand2Vec$ distances.

\subsection{Experiment: opening bid classification}

\begin{figure}
\centerline{\includegraphics[scale=0.7]{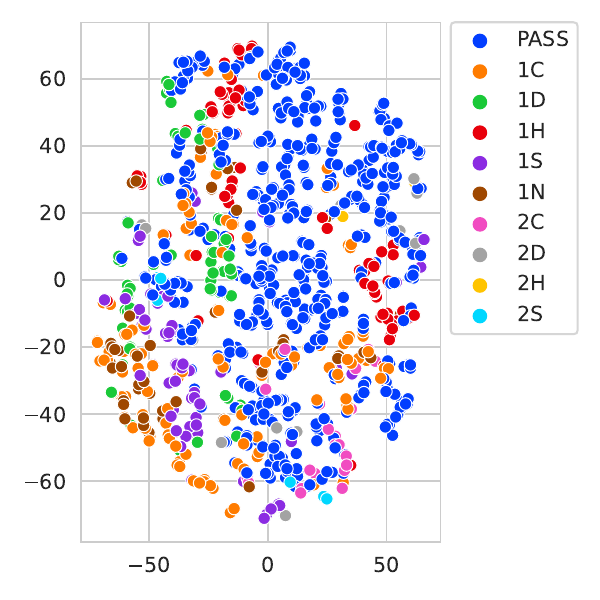}}
\caption{Opening bids (training set)} 
\label{fig:openings}
\end{figure}

Seven skilled\footnote{Members of the polish second bridge league} bridge players determined the best opening bid\footnote{Bidding system: Polish Club} for a set of 1213 randomly generated hands. Hands were split into training (1000) and test (213) sets and the training set was vectorized and embedded in a 2-dimensional space using t-SNE algorithm (Fig. \ref{fig:openings}). Frequency of the particular bids can be found in Table~\ref{tab:train_test}.

\begin{table*}
\centering
\caption{Opening bid frequency}
\label{tab:train_test}
 \begin{tabular}{|c|c|c|c|c|c|c|c|c|c|c|c|c|c|c|c|c|}
	\hline
	  & PASS & 1$\clubsuit$ & 1$\diamondsuit$ & 1$\heartsuit$ & 1$\spadesuit$ & 1NT & 2$\clubsuit$ & 2$\diamondsuit$ & 2$\heartsuit$ & 2$\spadesuit$ & 2NT & 3$\clubsuit$ & 3$\diamondsuit$ & 3$\heartsuit$ & 3$\spadesuit$ & 4$\heartsuit$ \\
	\hline
	Train & 549 & 127 & 64 & 76 & 64 & 51 & 22 & 14 & 3 & 6 & 3 & 11 & 4 & 1 & 3 & 2 \\
	\hline
	Test & 116 & 33 & 15 & 10 & 18 & 5 & 7 & 2 & 3 & 1 & 0 & 1 & 2 & 0 & 0 & 0 \\
	\hline
\end{tabular}
\end{table*}

We see (Fig. \ref{fig:openings}) that $BridgeHand2Vec$ dimensions are related to features used in "human" bidding systems:
\begin{itemize}
    \item Hands classified as the same opening are close to each other.
    \item Strong openings ($1\clubsuit$ and $1NT$) are close.
    \item Weak openings (e.g. $2\spadesuit$) are close to PASS and stronger openings with the same suit.
\end{itemize}

$BridgeHand2Vec$ vectorization of the training set was used to build a simple KNN classifier. Each hand from the test set was assigned the opening bid of the single nearest neighbour.

Additionally - four models which did not use the described vectorization - were prepared for comparison. One model is based on expert knowledge ($ExpertModel$), and three models are based on neural networks:
\begin{itemize}
    \item $RawModel$ - using only information about the player's cards;
    \item $HeuristicModel$ - using information about HCP, distribution of all suits, count of each figure - i.e. typical information, used by bridge players to determine an opening;
    \item $CombinedModel$ - using both of the above information.
\end{itemize}

$ExpertModel$ was developed according to the rules of the Polish Club bidding system. It determines the opening based on the number of points and cards in a given suit, using a set of conditional instructions. It does not consider other qualities of a hand, e.g. honor distribution, because it would be barely possible to incorporate.

Each network model consists of two multilayer perceptrons. One is used for determining the bid level, while the other is for the denomination. All networks have three hidden layers with the following number of neurons 8, 16, 8. The size of the input layer depends on the information the model uses and is 52, 9, and 61 neurons, respectively.

Table \ref{tab:model_acc} shows the accuracy of each model on the test set.

\begin{table}
    \centering
    \caption{Accuracy of the opening bid classifiers}
    \label{tab:model_acc}
    \begin{tabular}{c|c}
        Model & Accuracy\\\hline
        $BridgeHand2Vec$ & 71.36\%\\\hline
        $ExpertModel$ & 81.22\% \\\hline
        $RawModel$ & 69.48\% \\\hline
        $HeuristicModel$ & 78.87\%\\\hline
        $CombinedModel$ & 71.83\%
    \end{tabular}
\end{table}
$ExpertModel$ obtained the best result, but indeed it was not very impressive. This fact should be considered when evaluating other models. The best neural network model was $HeuristicModel$, and it is not unexpected because it uses the same heuristics as bridge players. $BridgeHand2Vec$'s accuracy was lower than that but higher than $RawModel$'s accuracy. Given that a very simple classifier was used (KNN), this result suggests that vectorization reflects the strength of the hand.

We should note that perfect accuracy is not possible on data generated by human players - selected bids are based not only on the rules defined by the bidding system but also on qualities such as e.g. unique values of the card\footnote{e.g. connected honours, lots of 10/9} or expected further auction. As these values are hard to formalize, different players fairly often select different bids for the same hand. We created a few hands that could be controversial:
\begin{itemize}
    \item Hand AK8765.Q76.J65.6 is too weak for a $1\spadesuit$ opening, but many players would nevertheless choose it because of the pretty spade suit. Opening 2$\diamondsuit$ (weak hand with six $\spadesuit$ or $\heartsuit$) is also an option.
    \item Despite having five spades, many players would open 1NT with AQ765.K5.QJ5.A65 because of distributed honours.
    \item Hand AJ102.KQ.65.QJ654 has a shape and strength matching $2\clubsuit$ opening, but many players would open $1\clubsuit$ instead, because of weak clubs.
    \item Hand AK65.AK65.K765.6 matches the $1\diamondsuit$ opening, but very good honours and four cards in both $\heartsuit$ and $\spadesuit$ would convince many players to open $1\clubsuit$ instead.
    
\end{itemize}
We checked if this controversy can be found in the vector representation. For each controversial hand, we found 4 nearest (in the vectorized space) hands in the training set. In each case, the nearest hands were classified into more than one opening (Table~\ref{tab:openings}). This sort of analysis is unavailable to programs based on the strict bidding rules - $BridgeHand2Vec$ representation is a step towards making robot bidding more human-like.

\begin{table}
\centering
\caption{Non obvious openings - similar hands}
\label{tab:openings}
 \begin{tabular}{c|c|c|c}
Hand & Nearest hands & Opening & Distance\\\hline
AK8765.Q76.J65.6 & AQJT6.J73.9864.J & PASS & 1.165\\
 & KQT73.A74.J976.Q & $1\spadesuit$ & 1.169\\
 & KJT97.Q963.A65.9 & PASS & 1.325\\
 & AQ8743.93.T32.T6 & 2$\diamondsuit$ & 1.369\\\hline
AQ765.K5.QJ5.A65 & AQ94.A86.QJ5.Q95 & 1NT & 1.099\\
 & AJT65.A4.K86.A73 & 1$\spadesuit$ & 1.141\\
 & KQT32.K9.AJ9.KT4 & 1NT & 1.204\\
& KT75.A3.KJ85.K64 & $1\clubsuit$ & 1.310\\\hline
 
AJT2.KQ.65.QJ654 & KT72.AK6.3.J8732 & 1$\clubsuit$ & 1.506\\
 & AJ72.KJT.3.AJ973 & 2$\clubsuit$ & 1.654\\
 & AJT2.KQ8.74.AQ32 & 1NT & 1.887\\
 & AQ5.QJ.852.AJ642 & 1$\clubsuit$ & 1.911\\\hline
 
AK65.AK65.K765.6 & AQT7.AQ83.QJ97.6 & 1$\diamondsuit$ & 0.861\\
 & KQJ9.A96.A742.62 & 1NT & 1.554\\
 & KQ93.A732.AT85.Q & 1$\diamondsuit$ & 1.617\\
  & AJ65.AQ62.J654.8 & 1$\diamondsuit$ & 1.742\\
 & AKJ9.KQ83.QJ8.A9 & 1$\clubsuit$ & 1.757\\ 
\end{tabular}
\end{table}

\subsection{RL learning for opening bid}

In order to check whether $BridgeHand2Vec$ might speed up the training process, we performed a simple reinforcement learning experiment.

The agent plays a game where they are supposed to guess the optimal contract, knowing only their hand. The reward for a single board is calculated as follows:
\begin{enumerate}
    \item Hands N and S are dealt. One is known to the agent, and the other is hidden.
    \item EW hands are dealt 10 times. For each of the complete deals, we calculate (using DDS \cite{Beling2017}) the best possible bridge score for NS\footnote{not vulnerable} and score for the guessed contract.
    \item The mean difference between the optimal score ($Sc$) and the obtained score constitutes the reward ($Rew$). Such rewards are always non-positive.
\end{enumerate}

The agent was trained using the Cross-Entropy Method. In each step, the agent processed 1000 different hands, of which elite (highest-reward) sessions were selected. Weights were updated to minimize the cross entropy loss between the probability distribution of the generated actions and the probability distribution of the decisions in the elite sessions.

In order to determine the impact of vectorization, we compare two agents. Both are neural networks with 3 hidden layers with 128 neurons and ReLu activation, but they work on different inputs. AgentBinary receives a 52-element binary vector, while AgentHand2Vec receives hand vectorization (8-element float vector). 

Table~\ref{tab:cem} shows a few sample decisions generated by both agents after 200 training steps and corresponding rewards and scores. AgentBinary learned to pass with every hand - this strategy is pretty reasonable (assuming no deeper knowledge about the game) because random bids usually result in negative scores. After the same number of steps, AgentHand2Vec learned to pass with weak hands and bid their best suit on level 1 with better hands. This is, on average, a better strategy than passing with every hand - we see that contracts are usually won (positive value in the column "AgentHand2Vec $Sc$").

\begin{table}
\centering
\caption{AgentBinary vs AgentHand2Vec decisions after 200 training steps}
\label{tab:cem}
 \begin{tabular}{c|c|c|c|c|c|c}
  & \multicolumn{3}{|c|}{AgentBinary} & \multicolumn{3}{|c}{AgentHand2Vec}\\\hline
Hand & Bid & $Rew$ & $Sc$ & Bid & $Rew$ & $Sc$\\\hline
95.AK632.KT2.AJ8 & PASS & -119 & 0  & 1$\heartsuit$ & 0 & 119\\\hline
AT84.AQJ65.A.432 & PASS & -456 & 0  & 1$\heartsuit$ & -250 & 206\\\hline
J74.Q93.K6.AT742 & PASS & -26 & 0     & 1$\clubsuit$ & 0 & 26\\\hline
J643.Q43.T973.AQ & PASS & -55 & 0  & 1$\diamondsuit$ & -175 & -120\\\hline
T732.95.AQ732.A3 & PASS & -36 & 0 & 1$\diamondsuit$ & 0 & 36\\\hline
83.JT.KQJ952.T92 & PASS & 0 & 0  & 1$\diamondsuit$ & -50 & -50\\\hline
J52.53.A942.KJ92 & PASS & -74 & 0     & 1$\diamondsuit$ & 0 & 74\\\hline
KQ86.96.Q754.J75 & PASS & 0 & 0   & 1$\diamondsuit$ & -12 & -12\\\hline
KQ43.A62.K.Q9762 & PASS & 0 & 0  & 1$\clubsuit$ & -130 & -130\\\hline
9742.KQT8.QJ73.K & PASS & -46 & 0     & 1$\diamondsuit$ & 0 & 46
\end{tabular}
\end{table}

\begin{figure}[H]
\centerline{\includegraphics[scale=0.8]{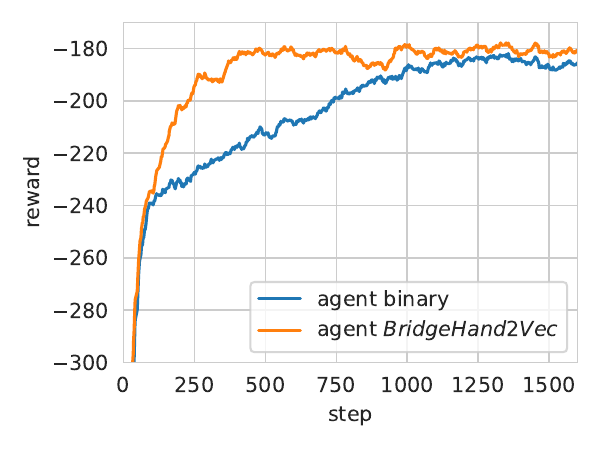}}
\caption{CEM learning speed up with vectorization} 
\label{fig:cem_learning}
\end{figure}

Fig.~\ref{fig:cem_learning} shows the average reward for both agents after a given number of steps. We see that, finally, both agents start to receive similar scores. Analysis of sample boards shows that they learned the same strategy (PASS with a weak hand, 1-level bid with a moderate hand, game bid with a strong hand). This might be the optimal strategy in this high-randomness game. The critical difference between the agents is the knowledge acquisition rate - AgentHand2Vec performs much better in the early training phase.

\section{CONCLUSION}
\label{sec:conclusion}

In this article, we propose $BridgeHand2Vec$ vectorization, where a bridge hand is represented as a vector of 8 floats. The neural network that uses this vectorization as the internal representation of input achieves SOTA results on the DDBP2 problem, which confirms that the representation captures features crucial for determining the strength of the hand. While this is itself an interesting result, we are more excited about the new possibilities that emerge from it:
\begin{itemize}
    \item Bridge hand algebra that exactly matches human intuitions. Statements like "hand X is closer to Y than to Z", or "hand X is between hands Y and Z", "hand X is a model hand for the given auction" based on $BridgeHand2Vec$ representation should match the same statements made by professional bridge players.
    \item More human-like bidding algorithms in bridge programs. Professional bridge players don't bid according to strict rules, but rather according to a very-hard-to-describe hand evaluation.
    \item Improved training of models that receive vectorization as an input.
\end{itemize}

\clearpage
\ack The authors would like to thank: Mateusz Kozłowski for the application used for the data collection from players, Rafał Przybysz, Konrad Paluszek, Jakub Andruszkiewicz, Łukasz Trendak (members of the team AZS UW Orły Warszawa) for data supply, and Karol Gałązka and Edward Sucharda for the consultations.

\bibliography{bib}
\end{document}